# Extracting Connected Concepts from Biomedical Texts using Fog Index

Rushdi Shams and Robert E. Mercer
*Department of Computer Science, University of Western Ontario, London, ON, N6A 5B7, Canada*
rshams@uwo.ca, mercer@csd.uwo.ca

**Abstract**
*In this paper, we establish Fog Index (FI) as a text filter to locate the sentences in texts that contain connected biomedical concepts of interest. To do so, we have used 24 random papers each containing four pairs of connected concepts. For each pair, we categorize sentences based on whether they contain both, any or none of the concepts. We then use FI to measure difficulty of the sentences of each category and find that sentences containing both of the concepts have low readability. We rank sentences of a text according to their FI and select 30 percent of the most difficult sentences. We use an association matrix to track the most frequent pairs of concepts in them. This matrix reports that the first filter produces some pairs that hold almost no connections. To remove these unwanted pairs, we use the Equally Weighted Harmonic Mean of their Positive Predictive Value (PPV) and Sensitivity as a second filter. Experimental results demonstrate the effectiveness of our method.*

**Key Words**- Computational Linguistics, Text mining, Connected Concepts, Fog Index, Text Readability.

## 1. Introduction

In recent years, extraction of connected biomedical concepts (i.e., disease, treatment, genes) from texts has drawn the attention of scientists interested in finding functional similarity (i.e., identification of genes involved in human diseases) [1]. Although benchmark research has reported successful methods to extract biomedical concepts [2] [3], they have rarely followed simple procedures. For example, Perez-Iratxeta *et al.* [4] could not relate diseases with gene functions from biomedical texts forthwith- they needed to apply a twofold intermediary process of connecting disease with chemical components and chemical components with gene functions. The key reason for this limitation of applying simple methods of extracting connected concepts from biomedical texts is manifold. While some researchers concentrated on number of occurrences of connected concepts in abstract of a paper [4] [5], others preferred to comb through either the full text [6] or pre-specified segments (i.e., Introduction, Methods, Results, and Discussion) [7]. Next to this, the connections can be very general (i.e., biochemical connections) or very specific (i.e., regulatory connections). Therefore, the demand of simple identification and extraction of biomedical or any other concepts from a scientific literature that maintain general or specific connections with one another is not met till to date. This situation suggests using improved yet simple computational method to identify and extract important, explicit and implicit connections from biomedical texts.

As text is highly structured by syntax and semantics of natural language, it is believed that such methods should involve these two features but several reports assert their complexity [8] [9]. Apart from this, Sherman [10] proposed that scientific literature is subject to statistical analysis and zeroed in on the importance of average sentence length. Gunning [11] practically demonstrated this important measure along with the number of complex words (i.e., words with three or more syllables) to assess the readability of text known as the Fog Index (FI), shorthand for the Gunning Fog Readability Index. It is now considered as a yardstick for readability assessment of books, novels, scientific literature and newspapers and even to detect online chatting bots [12].

In this paper, we report a simple novel statistical method to extract connected biomedical concepts from biomedical texts using FI. We statistically established FI as a text filter, experimenting on 24 random papers that describe four pairs of concepts: *Ischemia-Glutamate, Ataxia-Dehydrogenase, Hypogonadism-Gonadotropin,* and *Epilepsy-GABA*. Besides FI, our method also uses the equally weighted harmonic mean of the connections' Positive Predictive Value (PPV) and sensitivity as a second filter. While the prior concentrates on the important part of the text where the connections are stated, the latter assesses their representativeness. We selected sentences of a paper that are difficult to read and measured most frequent connected concepts present in them. With careful observations, we noticed that the first filter produces some *noisy* pairs of concepts that have a rather weak connection. To omit them, we calculated the PPV and sensitivity of each pair. Then, we filtered these pairs based on the equally weighted harmonic mean of their PPV and sensitivity.

In the remainder of this paper, we describe related work, illustrate the complete method, report and discuss experimental results, and draw conclusions.

## 2. Related work

New research trends in the biomedical field include the discovery of hidden connections in texts to form new hypotheses that can be explored further by conventional experimentation [6]. A series of investigations by Swanson [13] [14] showed that these hidden connections can lead us to new discoveries. He reported that fish oil leads to change in blood viscosity and red blood cell rigidity that helps prevent Raynaud's syndrome [13]. Later, investigative reports started to discover suggestions for clinical therapies and basic physiological linkages from bibliographically isolated texts. However, the working principles of Swanson's empirical research include the computational burden of full-text syntactic analysis and involve large literature databases like MEDLINE. Our work, though it does not generate hypotheses, can be a good means of finding implicit connections in texts using fewer computations (as it filters out texts according to their readability and does not operate syntactically) without involving literature repositories.

A handful of research work in semantic relation classification or extraction from bioscience texts depends on the proper identification of connections. Rosario and Hearst [15] concentrated on discovering connections between "treatment" and "disease". They reported 79.6 percent accuracy in blindly identifying concepts that fall into either of the categories and are somehow connected with one another. They used a MEDLINE-based neural network that addresses it to be intriguing yet complicated. A similar machine learning technique was applied by Frunza and Inkpen [8] to extract disease-treatment connections from texts. Their reported accuracy surpassed the results of Rosario and Hearst although their interest was limited to MEDLINE 2001 titles and abstracts. Their paper, like many other prominent work [16] [17], has a significant use of PPV and sensitivity to evaluate the mining technique. Contrary, we used these measures to evaluate the representativeness of the connected concepts.

Perez-Iratxeta *et al.* [4] proposed a massive framework to prioritize disease associated genes. Instead of looking into literature, they combined several isolated pieces of biomedical repositories like Medical Subject Heading (MeSH), Gene Ontology (GO), RefSeq database, and MEDLINE. They used both databases and ontology that have lack in communication with one another and thus experienced tedious and complex scoring methods and formidable number of intermediate stages. In our work, we decided to stick with texts only to remain simple yet capable of producing improved results.

Robert Gunning [11] first introduced Fog Index (FI) to measure semantic difficulties using average sentence length and polysyllabic words in his 1952 book *The Technique of Clear Writing*. We were motivated to apply FI as our text filter when we came across the experimental output carried by Duffy and Kabance [18]. They converted a passage with no more than two phrases into primer prose and applied FI to test its readability. They found the score well below the readability index (i.e., it was excessively easy to read). Their investigation on this phenomenon suggested that easy articles (in this case the primer prose) obscure the relationships and ideas as they emphasize each of them equally. In other words, difficult articles possess relationships and ideas and emphasize them in particular that yields low readability. We believe that if biomedical texts display similar attribute, then FI can be an appropriate measure to filter texts that bear associations of scientific interest.

## 3. Methodology

The work of Perez-Iratxeta *et al.* [4] lists pairs of connected concepts like disease-chemical components, chemical component-genes, and disease-genes. Among them, we considered four disease-chemical component pairs, namely *Ischemia-Glutamate, Ataxia-Dehydrogenase, Hypogonadism-Gonadotropin., and Epilepsy-GABA*. We collected 24 scientific papers (six for each pair of concepts) at random from several biomedical literature repositories. To work with the text only, we removed the title, affiliations, keywords, footnotes, figures, tables, acknowledgements, and references from the paper.

We considered each pair of concepts and a paper related to them. We classified its sentences into three sets: sentences containing both of the concepts, none of the concepts and any of the concepts. For example, the sentence "*Glutamate, which is potentially excito-toxic to brain neurons, is released excessively during ischemia*", will be put into the set of sentences containing both of the concepts *Ischemia* and *Glutamate,* as *Ischemia* and *Glutamate* are both present. Then, we applied Gunning's formula for FI (Eq. 1) to score the sentences of every set. It is noted that according to this formula, the lower the score of a sentence, the easier it is to read.

$$FI = 0.4 \times \left( \left( \frac{words}{sentences} \right) + 100 \times \left( \frac{complex\ words}{words} \right) \right) \quad (1)$$

It can be noted that according to Gunning, words that are polysyllabic (i.e., contain three or more syllables) are called $complex\ words$. Also, as we applied FI on every sentence, the value of $sentences$ is always 1.

We normalized this score by the paper's average number of syllables per word because readability score of long and short sentences varies due to the total number of syllables [11]. Eq. 2 provides the normalized FI ($FI'$) of the sentences in every set.

$$FI' = \frac{FI}{Average\ number\ of\ syllables\ per\ word} \quad (2)$$

Table 1 shows the $FI'$ calculated for the three sets of sentences for the 24 papers in groups related to the four pairs of connected concepts.

| Category | Ischemia-Glutamate | Remark | Ataxia-Dehydrogenase | Remark | Epilepsy-GABA | Remark | Hypogonadism-Gonadotropin | Remark |
|---|---|---|---|---|---|---|---|---|
| $FI'_{none}$ | 5.99 | Low | 5.77 | Low | 6.50 | Low | 5.58 | Low |
| $FI'_{both}$ | 8.26 | High | 7.23 | Medium | 7.24 | Medium | 10.29 | High |
| $FI'_{any}$ | 6.83 | Medium | 7.33 | High | 7.58 | High | 7.62 | Medium |

**Table 1. Normalized FI for three sets of sentences from 24 papers**

Table 1 shows that, for every pair of connected concepts, while the set of sentences containing any or both of the concepts displays either *High* or *Medium FI'*, the set of sentences containing none of them consistently possesses *Low* scores. This observation leads us to a decision that the sentences that are easier to read contain no connected concepts and therefore, we should look into low-readable sentences for hidden connections.

Now that we have FI as a functioning text filter, we need to define a way to determine the number of low-readable sentences to be considered for concept extraction. We ranked all sentences in a paper based on their FI score and sorted them in descending order (i.e., the most difficult sentences are at the top of the list). Then, in five chunks of 10 percent interval, we selected top 50 percent, 40 percent, 30 percent, 20 percent, and 10 percent of the sentences from this sorted list. For every chunk, we tagged these selected sentences with Genia Biomedical POS tagger [19], identified nouns in them and used an association matrix to record the frequency of connected concepts (i.e., number of occurrences of one noun with the other). For the sentence "*Glutamate, which is potentially excito-toxic to brain neurons, is released excessively during ischemia*", the connected concepts are *glutamate-brain, glutamate-neurons, glutamate-ischemia, brain-neurons, brain-ischemia,* and *neurons-ischemia*. From the output of the association matrix, we kept 20 most frequent for our experiment. We observed that some chunk $i$ contains new connections that are absent in chunk $i-1$ and vice versa. To find a threshold, we tracked number of connections revealed and connections missed by every chunk $i$ with respect to its previous chunk $i-1$. From Figure 1, we see that for the first chunk (50 percent of the sentences), all of the 20 most frequent connections are new. The number of new connections becomes steady in the third chunk (30 percent of the sentences) but reaches the extremes in the fourth and fifth. The results in Figure 1 are showed for six papers related to *Ischemia* and *Glutamate*. Similar experiments were conducted with other connected concepts and all of them showed that if we take less than 30 percent of the ranked sentences, the number of new concepts reach the extremes.

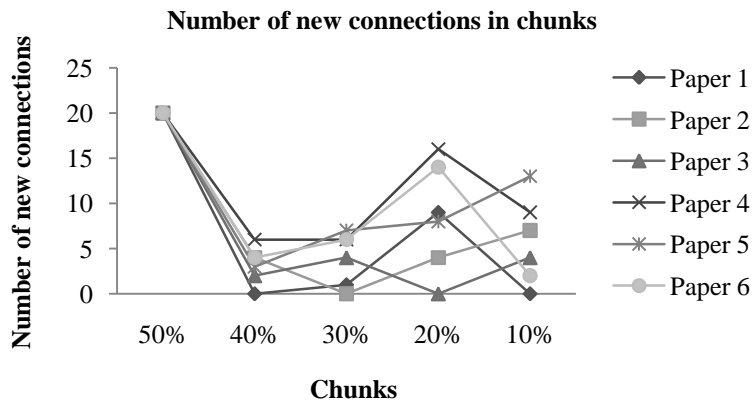

**Figure 1. Number of new connections in five chunks for six papers on Ischemia-Glutamate**

We recorded identical behavior for the number of connections dropped by every chunk. Figure 2 shows that as we start, the first chunk (10 percent of the sentences) does not miss any connection. The number of dropped connections becomes steady in the third chunk (30 percent of the sentences) but starts to reach the extremes in the fourth and fifth. Again, the results in Figure 2 are produced by six papers on *Ischemia* and *Glutamate*. We conducted similar experiments with other connected concepts. All of them showed that if we take less than 30 percent of the ranked sentences, the number of dropped connections reach the extremes.

These two observations indicate that the degree of concepts connected with each other is conserved if we take 30 percent of the low-readable sentences. Similar results are obtained for the three other pairs of concepts.

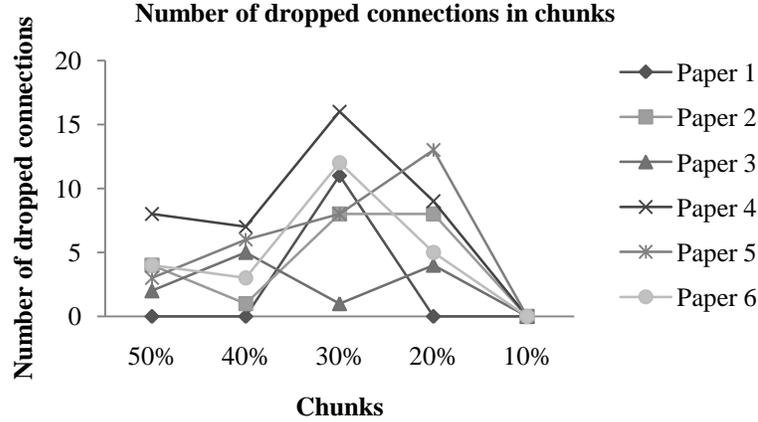

**Figure 2. Number of dropped connections in five chunks for six papers on Ischemia and Glutamate**

Provided the threshold, Table 2 shows the 20 most frequent connected concepts found in a paper on *Ischemia-Glutamate* where the connections are ranked according to their frequency. For each pair shown in Table 2, we extracted sentences from the paper that contain both of the concepts. These sentences are fed to Unified Medical Language System (UMLS) semantic relation network [20] to find out the semantic relations between the concepts. Surprisingly, we found that among the 20 connected concepts, only nine have textual semantic connections (Levels-Glutamate, Ischemia-Glutamate, Levels-Increase, Increase-Glutamate, 10min-Ischemia, Glutamate-Experiment, Glutamate-Neurons, Glutamate-CA4, and Ischemia-5min).

| Rank | Connected Concepts | Frequency | Semantic Connection | Rank | Connected Concepts | Frequency | Semantic Connection |
|---|---|---|---|---|---|---|---|
| 1 | Levels-Glutamate | 48 | Yes | 8 | 10min-Ischemia | 17 | Yes |
| 2 | Ischemia-Glutamate | 37 | Yes | 9 | Glutamate-Neurons | 16 | Yes |
| 3 | Levels-Ischemia | 33 | No | 9 | Levels-5min | 16 | No |
| 4 | Levels-10min | 22 | No | 9 | Glutamate-Experiment | 16 | Yes |
| 4 | 10min-Glutamate | 22 | No | 10 | Levels-Neurons | 15 | No |
| 5 | Levels-Half | 21 | No | 10 | Glutamate-CA4 | 15 | Yes |
| 5 | Levels-Increase | 21 | Yes | 10 | Levels-CA4 | 15 | No |
| 6 | Increase-Glutamate | 20 | Yes | 11 | Ischemia-5min | 14 | Yes |
| 6 | Glutamate-5min | 20 | No | 11 | Levels-Pretreatment | 14 | No |
| 7 | Half-Glutamate | 19 | No | 11 | Levels-Experiment | 14 | No |

**Table 2. Most frequent connected concepts for a paper on Ischemia and Glutamate**

So, FI, as a text filter, brings in some text that contains most frequent connected concepts but some of the concepts lack representativeness (i.e., they do not hold any connection). It urged us to provide a means by which we can filter out these *noisy* pairs of concepts. We observed as we collected texts at random, there is the possibility that the pairs we considered for experimentation may never co-occur in a sentence which indicates that our data set is imbalanced. So, we used the equally weighted harmonic mean of the PPV and sensitivity of the pairs of concepts provided by FI to evaluate their representativeness as it is a great evaluation metric for imbalanced dataset [8].

PPV is the percentage of correctly predicted connections and sensitivity represents the percentage of connections identified as relevant by our system. To measure the PPV and sensitivity of every pair of concepts, we first considered the set of sentences filtered by FI and counted the number. This is the total number of results returned by our system ($R$) that comprises the number of True Positives ($TP$) and False Positives ($FP$). Then, we take a pair depicted in Table 2, searched the paper, and developed a second set of sentences that contain both of its concepts. The number of sentences in this set is the number of results that should have been returned by our system ($S$) and comprises the number of True Positives ($TP$) and False Negatives ($FN$). Finally, we counted the number of sentences that are present in both sets- which is the number of $TP$s returned by our system. Afterwards, $FP$ is obtained by subtracting $TP$ from $R$ and $FN$ is obtained by subtracting $TP$ from $S$. So, the PPV of every pair of connected concepts is $\frac{TP}{TP+FP}$ and the sensitivity of every pair of connected concepts is $\frac{TP}{TP+FN}$. We then applied the formula in Eq. 3 to determine the equally weighted harmonic mean for the given pair of concepts. In this way, we measured this mean for every pair of concepts in Table 2.

*Harmonic Mean of PPV and Sensitivity*
$$= 2 \times \left(\frac{PPV \times Sensitivity}{PPV + Sensitivity}\right) \quad (3)$$

When we finished measuring this mean for all of the pairs, we re-ranked them and considered the first 10 pairs of concepts. These 10 pairs of connected concepts are said to be the representative connected concepts of the paper.

We followed the similar procedure to evaluate the representativeness of pairs of concepts for the rest of the three connected concepts.

We measured the accuracy of every connected pair by using Eq. 4 as well-
$$Accuracy = \frac{TP+TN}{TP+TN+FP+FN} \quad (4)$$

Where $TN$ is the number of True Negatives and can be found by subtracting $(TP + FP + FN)$ from total number of sentences in a text. But we found that in the case of accuracy, the connections are not distinguishable according to their ranks.

## 4. Results and Discussions

Table 3 lists the 10 connected concepts for a paper on Ischemia and Glutamate among which seven pairs of concepts are reported as semantically connected by UMLS.

| Rank | Connected Concepts | Harmonic Mean | Semantic Connection |
|---|---|---|---|
| 1 | Ischemia-Glutamate | 51.85 | Yes |
| 2 | Levels-Ischemia | 43.47 | No |
| 3 | Levels-Glutamate | 41.66 | Yes |
| 4 | Glutamate-Neurons | 39.02 | Yes |
| 5 | 10min-Ischemia | 37.50 | Yes |
| 6 | Glutamate-CA4 | 35.89 | Yes |
| 7 | Increase-Glutamate | 32.55 | Yes |
| 8 | 10min-Glutamate | 31.81 | No |
| 9 | Ischemia-5min | 31.57 | Yes |
| 9 | Glutamate-5min | 31.57 | No |

**Table 3. Connected concepts for a paper on Ischemia and Glutamate**

Table 4 shows the 10 connected concepts for a paper on Ataxia and Dehydrogenase, eight of which are semantically connected in the UMLS semantic relation network. Our observation of this domain reveals that PDHC (Pyruvate Dehydrogenase Complex) is manifested in Ataxia patients, especially those suffering from Friedreich's Ataxia. So, the relations among Friedreich, Ataxia and PDHC are vividly represented in the list.

| Rank | Connected Concepts | Harmonic Mean | Semantic Connection |
|---|---|---|---|
| 1 | Friedreich-Ataxia | 59.25 | Yes |
| 2 | PDHC-Ataxia | 56.00 | Yes |
| 3 | Activity-Friedreich | 43.47 | Yes |
| 3 | Patients-Ataxia | 43.47 | Yes |
| 3 | Activity-Ataxia | 43.47 | Yes |
| 3 | PDHC-Friedreich | 43.47 | Yes |
| 4 | Preparations-Ataxia | 40.00 | No |
| 4 | Preparations-Friedreich | 40.00 | No |
| 5 | Pyruvate-Ataxia | 38.09 | Yes |
| 6 | Patients-Friedreich | 36.36 | Yes |

**Table 4. Connected concepts for a paper on Ataxia and Dehydrogenase**

Table 5 lists the 10 connected concepts for a paper on Hypogonadism and Gonadotropin. According to the UMLS semantic relation network, eight of these pairs are semantically related.

| Rank | Connected Concepts | Harmonic Mean | Semantic Connection |
|---|---|---|---|
| 1 | AAS-Treatment | 29.41 | Yes |
| 2 | Use-AAS | 21.62 | No |
| 3 | AAS-Testosterone | 18.46 | Yes |
| 4 | Gonadotropin-Treatment | 18.18 | Yes |
| 5 | Testosterone-Treatment | 14.92 | Yes |
| 6 | Levels-Testosterone | 14.49 | Yes |
| 7 | AAS-Conditions | 12.90 | Yes |
| 7 | Treatment-HCG | 12.90 | Yes |
| 7 | Replacement-Therapy | 12.90 | No |
| 7 | Treatment-Therapy | 12.90 | Yes |

**Table 5. Connected concepts for a paper on Hypogonadism and Gonadotropin**

Steroids have significant effects on diseases like Hypogonadism, where release of testosterone plays an important role. Therefore, the connection between AAS (Anabolic Androgenic Steroid) that induces Hypogonadism and Testosterone is present in the list.

Table 6 displays the connected concepts present in a paper on Epilepsy and GABA.

| Rank | Connected Concepts | Harmonic Mean | Semantic Connection |
|---|---|---|---|
| 1 | Inhibition-GABA | 26.08 | Yes |
| 2 | GABA-Synapse | 20.25 | Yes |
| 3 | Neurons-Synapse | 14.70 | Yes |
| 4 | Inhibition-Hippocampus | 12.30 | Yes |
| 5 | Synapse-Change | 9.37 | Yes |
| 6 | Neurons-GABA | 8.00 | Yes |
| 7 | Properties-GABA | 6.45 | Yes |
| 7 | GABA-Change | 6.45 | No |
| 8 | GABA-Number | 6.34 | No |
| 9 | Cl-Gradient | 3.33 | Yes |

**Table 6. Connected concepts for a paper on Epilepsy and GABA**

Epilepsy is a neuronal disease that causes inhibition and significantly affects neuronal structure like the Hippocampus. In the list, we find eight concepts that are semantically related according to UMLS.

## 5. Conclusions

In this paper, we report on the extraction of connected concepts from biomedical texts by assessing text readability. The readability of text is determined by a metric called Fog Index (FI). We curated 24 random papers by using four pairs of connected concepts as keywords and applied FI on them. Experimental results showed that sentences display low readability if they contain connected concepts. We selected 30 percent of the most difficult-to-read sentences, and used an association matrix to track the most frequent pairs of concepts in them. To remove those pairs of concepts that have a rather weak connection, we used the equally weighted harmonic mean of their positive predictive value and sensitivity as a second ranking filter. The results are supported by finding almost all of the extracted concepts semantically connected by the UMLS semantic relation network.